\documentclass{bmvc2k}

\usepackage[mathscr]{euscript}
\usepackage{multirow}
\usepackage{times}
\usepackage{epsfig}
\usepackage{graphicx}
\usepackage{amsfonts}
\usepackage{amsmath}
\usepackage{amssymb}
\usepackage{comment}
\usepackage{color}
\usepackage{xcolor}
\usepackage{multirow}
\usepackage{balance}
\usepackage{enumitem}
\usepackage{adjustbox}
\usepackage{color, colortbl}

\definecolor{LightCyan}{rgb}{0.88,0.88,1}
\definecolor{Gray}{gray}{0.9}

\usepackage{hhline}
\usepackage{makecell}

\usepackage{algpseudocode}
\usepackage[linesnumbered,ruled,vlined]{algorithm2e}
\usepackage{wrapfig,lipsum,booktabs}

\def\net{\emph{FiFA}}

\title{Fiducial Focus Augmentation for Facial Landmark Detection}

\addauthor{Purbayan Kar}{purbayan.kar@sony.com}{1}
\addauthor{Vishal Chudasama}{vishal.chudasama1@sony.com}{1}
\addauthor{Naoyuki Onoe}{naoyuki.onoe@sony.com}{1}
\addauthor{Pankaj Wasnik$^{\dagger}$}{pankaj.wasnik@sony.com}{1}

\addauthor{Vineeth Balasubramanian}{vineethnb@cse.iith.ac.in}{2}

\addinstitution{
 Sony Research India,\\
 Bangalore, India
}
\addinstitution{
 Indian Institute of Technology, \\
 Hyderabad, India
}

\runninghead{Purbayan et al.}{Fiducial Focus Augmentation for Landmark Detection}

\begin{document}

\maketitle

\begin{abstract}
Deep learning methods have led to significant improvements in the performance on the facial landmark detection (FLD) task. However, detecting landmarks in challenging settings, such as head pose changes, exaggerated expressions, or uneven illumination, continue to remain a challenge due to high variability and insufficient samples. This inadequacy can be attributed to the model's inability to effectively acquire appropriate facial structure information from the input images. To address this, we propose a novel image augmentation technique specifically designed for the FLD task to enhance the model's understanding of facial structures. To effectively utilize the newly proposed augmentation technique, we employ a Siamese architecture-based training mechanism with a Deep Canonical Correlation Analysis (DCCA)-based loss to achieve collective learning of high-level feature representations from two different views of the input images. Furthermore, we employ a Transformer + CNN-based network with a custom hourglass module as the robust backbone for the Siamese framework. Extensive experiments show that our approach outperforms multiple state-of-the-art approaches across various benchmark datasets.
\end{abstract}

\section{Introduction}
\label{sec:intro}
Facial Landmark Detection (FLD) aims to detect coordinates of the predefined landmarks on given facial image.  The rich geometric information provided by landmarks with distinct semantic significance, such as eye corner, nose tip, or jawline, can be helpful in various tasks like 3D face reconstruction \cite{reconstruction6,reconstruction7,reconstruction8}, face identification \cite{recognition1,recognition2,recognition3}, emotion recognition \cite{emotion1,emotion2,emotion3}, and face morphing \cite{facemorphing}. 
Several FLD algorithms, based either on coordinate regression \cite{coord1,coord2,coord3,coord4,coord5,coord6} or heatmap regression \cite{farl, shfan, propnet,dtld,picassonet,hih}, have emerged in recent years with promising performance on various datasets. However, landmark detection still remains challenging task due the high variability in poses, lighting and expressions.
Despite the various existing FLD methodologies, none have focused on robust image augmentation techniques to solve these challenges. This study illustrates that meticulously designed image augmentations can considerably enhance the FLD performance. 




\begin{figure}[t!]
    \centering
    \includegraphics[width = 0.83\linewidth, height = 0.295\textheight]{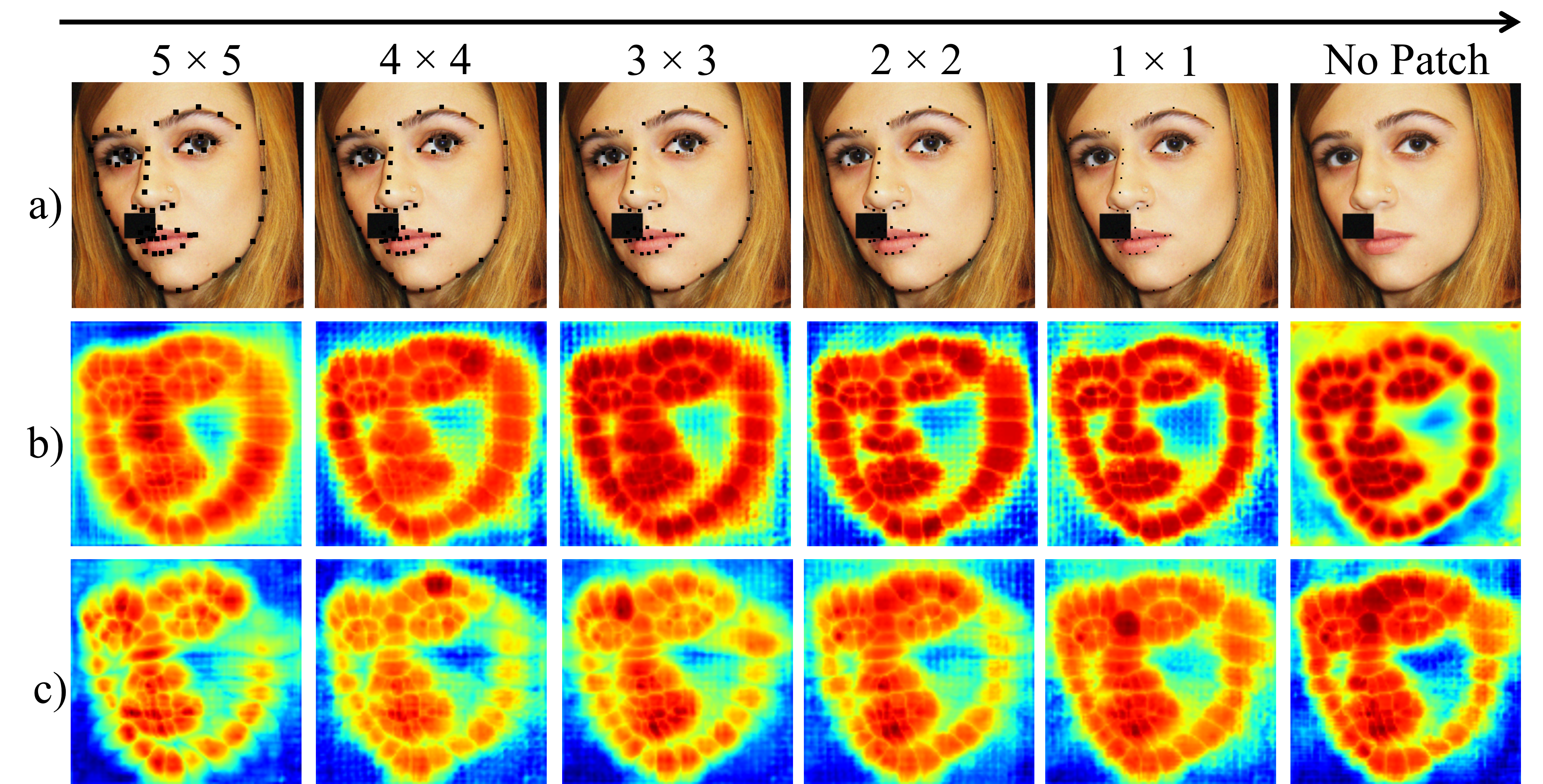}
    \caption{\small \textbf{Illustration of the proposed Fiducial Focus Augmentation (\net).} In row (a), 5$\times$5 black patches are created around the landmark joints (along with other standard augmentations) in the initial epochs and reduced over the epochs. Rows (b) and (c) show corresponding GradCAM-based saliency maps of the network's last layer with and without \net, respectively. It is clearly seen that activations are more prominent around the desired landmarks when \net\ is used as additional augmentation.}
    \label{fig:intro}
    \vspace{-1em}
\end{figure}

But why do sophisticated deep neural network (DNN) architectures struggle to detect landmarks accurately in challenging scenarios? The reason is that the DNN is unable to learn the facial structure information as accurately as required. If a DNN model can accurately capture features that extract a facial structure, it can predict the landmarks more accurately even from obscured facial regions, like occluded areas. To learn facial structures effectively, we propose new augmentation technique called Fiducial Focus Augmentation (\net), which leverages the ground truth landmark coordinates as an inductive bias for facial structure.
To this end, we introduce $n\times n$ black patches around the landmark locations in the training images, gradually reducing them over the epoch and then removing completely for the rest of the training, as illustrated in Fig \ref{fig:intro}. Since the patches cover key semantic regions of the face, e.g., eyes, nose, lips and jawline, when the model learns to predict these patches, it is able to learn the entire facial structure significantly better, as compared to an architecture without this inductive bias. One could view this augmentation technique as similar to Curriculum Learning (CL) \cite{curriculum}, a strategy that trains a machine learning model from simpler data to more difficult data, mimicking the meaningful order found in human-designed learning curricula.



Drawing inspiration from \cite{shfan}, we leverage the Siamese architecture to acquire a comprehensive understanding of reliable landmark predictions across various image augmentations. 
However, our method employs Deep Canonical Correlation Analysis (DCCA) \cite{dcca} as loss function in Siamese architecture to amplify the efficacy of the learning process between distinctively augmented views.
This loss function assists in the extraction of features that are correlated across views, while simultaneously eliminating uncorrelated noise. 
To design a robust backbone for the Siamese architecture, we adopt Vision Transformer (ViT) \cite{ViT}. We further improved its performance and efficiency by incorporating a Convolutional Neural Network (CNN)-based hourglass module in-between the transformer layers of the ViT. Modern CNNs are usually considered to be shift-invariant; we hence use an Anti-aliased CNN \cite{antialiasing} inside the hourglass module to leverage this benefit. 
We summarize the contributions of this paper as follows.
\vspace{-0.5em}
\begin{itemize}[leftmargin=*]
\setlength\itemsep{-0.4em}
    \item To the best of our knowledge, this is the first effort in literature to propose a new patch-based augmentation technique for FLD task to learn facial semantic structures effectively.
    \item We employ a Siamese-based training scheme utilising DCCA loss between feature representations of two different views of the same image, that enforces consistent predictions of the landmark for the two views. To incorporate virtues of both a Transformer and a CNN, we design a robust Transformer + CNN-based backbone in our proposed framework.
    \item We performed extensive experiments on various benchmark datasets showing significant improvements over prior work. We also conducted ablation studies on our framework components and additional empirical analysis to study the usefulness of the proposed method.
\end{itemize}

\section{Related Works}
\label{sec:related}
Earlier efforts on FLD task, especially those in recent years, can broadly be categorized into network architecture enhancements for heatmap generation and loss function improvements. 

\vspace{1pt}
\noindent \textbf{Network architecture enhancements:} Coordinate regression-based methods \cite{coord1, coord2, coord3, coord4, coord5, coord6} directly perform regression on landmark coordinate vectors through a fully connected output layer that disregards the spatial correlations of features and results in limited accuracy of landmark detection. On the other hand, heatmap regression-based methods \cite{shfan, bulat2017far, farl, adnet, propnet, hih, slpt, picassonet, dtld} predict landmark coordinates by creating heatmaps. By doing so, they effectively maintain the original spatial relationships between pixels and achieve promising landmark detection accuracy. Therefore, heatmap regression has become the de facto choice for the FLD task in modern times. In \cite{bulat2017far}, Bulat \emph{et al.} proposed an encoder-decoder based framework with heatmap regression for FLD. Their network incorporates hourglass and hierarchical blocks. Several research works \cite{sun2019high, wang2020deep, xiao2018simple} have been published based on the ResNet \cite{he2016deep} architecture and modify their network for dense pixel-wise landmark predictions. Recently, the Vision Transformer (ViT) \cite{ViT} has been incorporated in FLD task by Zhang \emph{et al.} \cite{farl} and has produced remarkable results. In our proposed framework, we also use ViT as the backbone network and improve its performance by introducing CNN layers in between transformer layers. This allows us to combine the best of both designs.

\vspace{1pt}
\noindent \textbf{Loss function improvements:} A pixel-wise $L2$ or $L1$ loss is the conventional loss generally applied to heatmap regression-based methods \cite{heatmap1, heatmap2, heatmap3, heatmap4, heatmap5}. To emphasize the importance of tiny and medium range errors during the training process, Feng \emph{et al}. \cite{wing} introduced the Wing loss, which modifies the L1 loss by using a logarithmic function to amplify the impact of errors within a specific range. Additionally, Wang \emph{et al}. \cite{wang2019adaptive} developed the Adaptive Wing Loss, which can adjust its curvature based on the ground truth pixels. In \cite{kumar2020luvli}, Kumar \emph{et al}. proposed the LUVLi loss that optimizes the position of the keypoints, the uncertainty, and the likelihood of visibility. Recently, the authors from \cite{propnet} proposed the Focal Wing Loss, which is used to mine and emphasize difficult samples under in-the-wild conditions. 

In this work, we use the standard Binary Cross Entropy (BCE) and $L2$ losses for heatmap and coordinate regression, respectively. We however employ the DCCA loss \cite{dcca} which suits our framework and has never been used before for the FLD task. These simple losses help the proposed framework set a new benchmark. Our study of literature revealed that well-designed image augmentations are largely ignored for the FLD task. This paper attends to this very issue and introduces a new augmentation technique called \net\ that accounts for our impressive results.

\begin{figure}[t!]
    \centering
    \includegraphics[width = 0.999\linewidth]{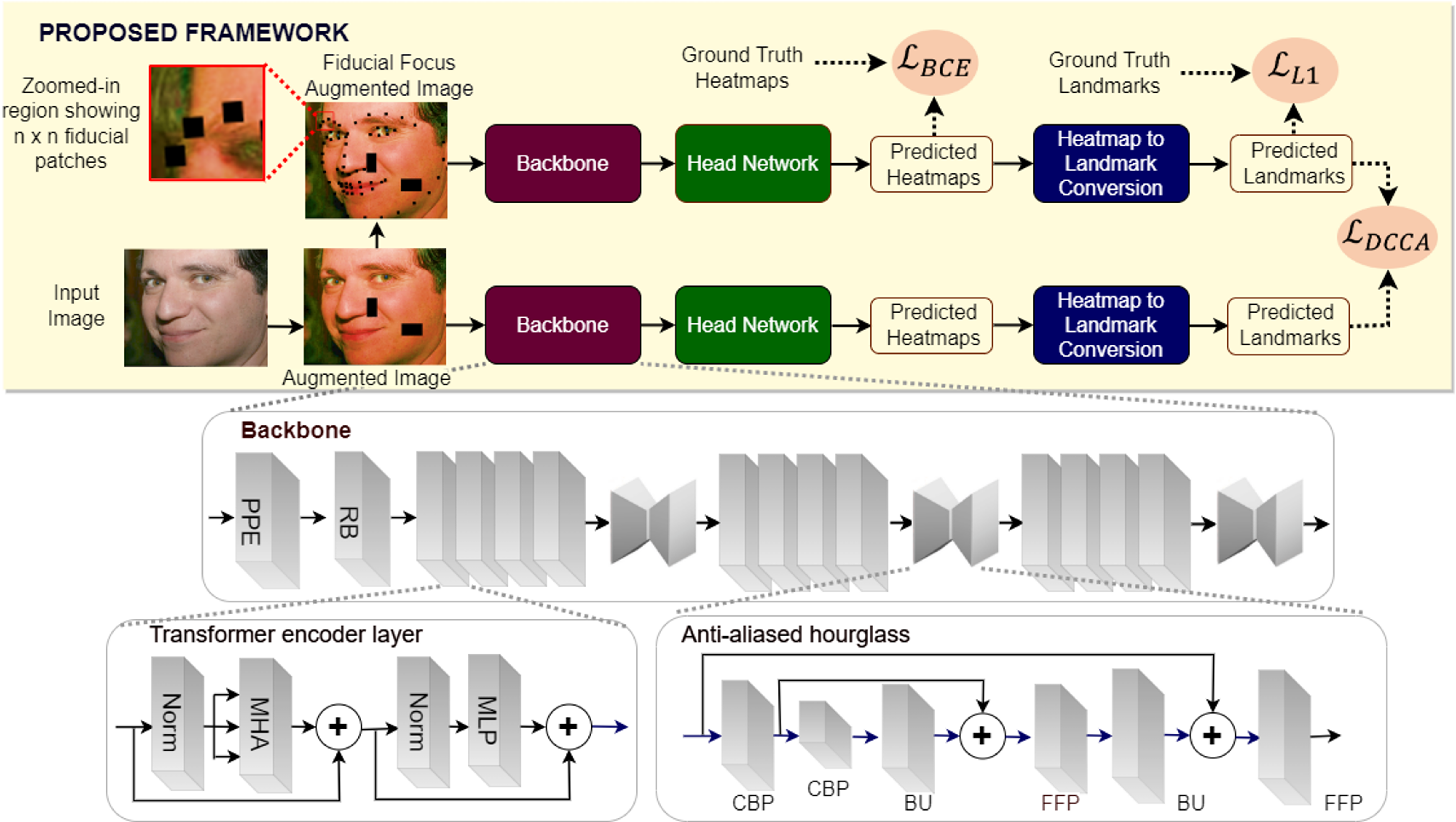}
    \vspace{-1em}
    \caption{\small \textbf{An overview of the proposed Siamese-based framework}. PPE = Patch + Position Embeddings; RB = Residual Block; MHA = Multi-Head Attention, MLP = Multi-Layer Perceptron; CBP = Convolution+BlurPool; BU = Bilinear Upsampling; FFP = FF-Parser.
    }
    \label{fig:network}
\end{figure}


\section{Proposed Framework}
\label{sec:method}
\subsection{Problem Statement \& Notations}

Given an input image $I$, FLD aims to detect $\{x,y\}\in \mathbb{R}^{k\times 2}$, the coordinates of $K$ predefined landmarks. To this end, we propose a heatmap-based approach to regress the facial landmarks. During training, it encodes the target ground truth coordinates as a series of $k$ heatmaps with a 2D Gaussian curve centered on them:
\vspace{-0.4em}
\begin{equation}\label{eq:eq1}
    \Psi_{i,j,k} = \frac{1}{2\pi\sigma^{2}}e^{-\frac{1}{2\sigma^{2}}\left[(i-\bar{x}_{k})^{2}+(j-\bar{y}_{k})^{2}\right]} 
\end{equation}
where $x_{k}$ and $y_{k}$ are the spatial coordinates of the $k^{th}$ point, while $\bar{x}_{k}$ and $\bar{y}_{k}$ are their scaled, quantized version obtained by scaling factor $s$ and rounding operator $\lfloor\cdot\rceil$, i.e.
\vspace{-0.5em}
\begin{equation}
    (\bar{x}_{k},\bar{y}_{k}) = (\lfloor\frac{1}{s}x_{k}\rceil,\lfloor\frac{1}{s}y_{k}\rceil) \label{eq:eq2}
\end{equation}
 
As shown in Eq. \eqref{eq:eq1}, we use a Gaussian with variance $\sigma$ around each coordinate from $\{x,y\}$
to generate the corresponding heatmap $\mathbb{H} \in \mathbb{R}^{k\times W\times H}$. Finally, the pixels with maximum intensity of the heatmap $\mathbb{H}$ are selected to get the final $K$ landmarks in the FLD task. 

To attain precise facial landmarks, we propose a novel augmentation technique called Fiducial Focus Augmentation (\net) that helps the network to learn facial structures in the provided images, along with a Siamese network with a robust backbone and the DCCA loss to ensure consistent predictions between different augmented views. 
Detailed explanations of these modules are provided in the subsequent subsections.
\subsection{Fiducial Focus Augmentation}
We seek to explore the potential of carefully designed image augmentations for the FLD task in this section. To this end, we propose an augmentation $f_A$ for input training images, where  $f_A = f_{A_2} \circ f_{A_1}$. Here, $f_{A_1}$ can be any standard image augmentations used in the FLD task \cite{farl, shfan, propnet, slpt, picassonet} and $f_{A_2}$ is the proposed Fiducial Focus Augmentation (\net).

First, we take the original input image $I$ and apply standard image augmentation $f_{A_1}$ to get the augmented image ($I'$). Mathematically, this can be expressed as:
\vspace{-0.5em}
\begin{equation}
    I' = f_{A_1} \otimes I.
\end{equation} 
To get the final augmented image $I''$, $I'$ is passed through the proposed augmentation operation i.e., $f_{A_2}$ (as descibed in Alg \ref{Algorithm}), i.e.
\vspace{-0.2em}
\begin{equation}
    I'' = f_{A_2} \otimes I' = \hat{I} \otimes I' = \hat{I} \otimes (f_{A_1} \otimes I).
\end{equation}
Here, we aim to incorporate the available facial structure ground truth information into the augmented image, $I'$ in order to aptly utilize the underlying facial structure. To achieve this, we construct black square patches of dimensions $h_{f}\times w_{f}$, where $h_{f},w_{f} \in \{1,\cdots,n\}$ while retaining the landmarks as the intersection points of the two diagonals of the square patches (see Figure \ref{fig:intro} (a)). These patches comprise of four coordinates which can be expressed as:
\begin{equation}
    \{(x_{i}-w_{f},y_{i}+h_{f}),(x_{i}+w_{f},y_{i}+h_{f}),(x_{i}+w_{f},y_{i}-h_{f}),(x_{i}-w_{f},y_{i}-h_{f})\}\ \forall\ \{x_{i},y_{i}\} \in L.
\end{equation}
Here, we start with a bigger patch size of $n\times n$ for a certain number of epoch intervals $\mathcal{E}$. After every such interval, we reduce the patch size by 1 pixel and eventually, these patches are removed from the images and rest of the training goes on with augmentation $f_{A_1}$ only. So the final augmented image is (where $T_n$ is the total number of epochs):
\vspace{-0.6em}
\begin{equation}
\begin{cases}
I'' &\text{when epoch no. $\leq n\cdot \mathcal{E}$}\\
I' &\text{when $n\cdot \mathcal{E}$ < epoch no. $\leq T_n$}.
\end{cases}
\end{equation}
\vspace{-0.6em}

\begin{algorithm}[H]
\footnotesize 
\SetKwInput{KwInput}{Initialize}                
\SetKwInput{KwOutput}{Output}              
\DontPrintSemicolon
\KwInput{\ \ $I'$: Augmented Image, where $I' = f_{A_1} \otimes I$, \\ 
\quad \qquad\quad\ $L_{n}$: Number of landmarks in $I$,\\ 
\quad \qquad\quad\ \, $L$: Set of $L_{n}$ landmarks, where $L = \{(x_{i},y_{i})\}$, $i \in \{1,...,L_{n}\}$, \\
\qquad \ \, $h_{f},w_{f}$: Height and width of the patches $(S)$, where $h_{f},w_{f} \in \{n,...,1\}$, \\
\qquad\quad\quad \ $I_{in}$: Pixel intensity of $S$, where $I_{in}=(0,0,0)$, \\
\qquad\qquad \ \, $\mathcal{E}$: Epoch interval, where $ \mathcal{E} \in \{1,...,n\} \land n <$ Total number of epochs $(T_n)$\\ 
\quad\qquad\quad\ $T_n:$ Total number of epochs $=\sum_{i=1}^n\mathcal{E}_i + w$, where $w \in \mathbb{W}$ \\
\textbf{Procedure:}\\
\qquad \textbf{for} $i$ in range $T_n$ \textbf{do} \\
\qquad \qquad  $k \leftarrow \lfloor i / \mathcal{E} \rfloor$      \\
\qquad \qquad $h_f, w_f \leftarrow |n - k|$  \\
\qquad \qquad \textbf{for} $j$ in range $L_n$ \textbf{do} \\
\qquad \qquad \qquad $C \leftarrow \{(x_{j}-w_{f}/2,y_{j}+h_{f}/2),(x_{j}+w_{f}/2,y_{j}+h_{f}/2),(x_{j}+w_{f}/2,y_{j}-h_{f}/2),(x_{j}-w_{f}/2,y_{j}-h_{f}/2)\}$   \\
\qquad \qquad \qquad Create patch $S$ with $C$ of $I_{in}$  \\
\qquad \qquad \qquad $I' \leftarrow S \otimes I'$ \\
\qquad \qquad \textbf{end for} \\
\qquad \textbf{end for} \\
\qquad $\hat{I} \leftarrow I'$ \\
\qquad \textbf{return} $\hat{I}$
}
\caption{\small Fiducial Focus Augmentation $(f_{A_2})$}\label{Algorithm}
\end{algorithm}

The proposed \net\ helps the backbone network learn the underlying facial structure and address difficult test samples, since the patches cover the entire face uniformly over the different joints (eyes, lips, nose and jawline). At the beginning of training, the model is exposed to larger patches as low-confidence regions to concentrate on the joints and eventually, as the model learns progressively with each epoch, smaller patches are introduced as high-confidence regions around the joints. When the patches are removed completely, the model tries to predict the joints with the inductive bias provided by earlier training steps in our augmentation process. Since the patches can be used with any facial variations (such as pose or expression), their integration into the images as augmentations enables the model to learn the inherent facial structures.

\subsection{Matching Two Views}
Earlier work on the task of FLD has seen limited exploration of Siamese architecture-based training, with the exception of \cite{shfan}. In this paper, we propose a Siamese architecture-based framework as illustrated in Fig.~\ref{fig:network}. The network $f$ takes the two input images $I'$ and $I''$ generated using two different augmentations $f_{A_1}$ and $f_{A}$. This training scheme using augmentations holds a notable advantage, as CNNs may not be invariant under arbitrary affine transformations. Therefore, even minor variations within the input space may produce significant changes in the output. By optimizing jointly using the Siamese architecture and combining the two predictions, we enhance the robustness and consistency of the predictions (under such variations).

To maximize the correlation between two different augmented views, we employ the Deep Canonical Correlation Analysis (DCCA) loss \cite{dcca} between the high-level representation mappings $f_{1}(I')$ and $f_{2}(I'')$, where $f_{1} = f_{2} = f$. The correlation between these two mappings can be expressed as below:
\vspace{-0.7em}
\begin{equation}
    corr(f_{1}(I'),f_{2}(I''))=\frac{cov(f_{1}(I'),f_{2}(I''))}{\sqrt{var(f_{1}(I'))\cdot var(f_{2}(I''))}}.
\end{equation}
\vspace{-0.25em}
The DCCA loss (i.e., $\mathcal{L}_{DCCA}$) is then computed as:
\vspace{-0.5em}
\begin{equation}
    \mathcal{L}_{DCCA} = - corr(f_{1}(I'),f_{2}(I'')).
\end{equation}

\vspace{-0.25em}
The use of DCCA loss presents three key advantages: (i) correlated representations partially reconstruct the information in the second view, when it is unavailable; (ii) it has potential to eliminate noise that is uncorrelated across the two views; and (iii) if $f_1, f_2$ capture features that are correlated across the views, they may represent latent aspects of the face. This, in turn helps the backbone network in capturing the facial structure in the images.
\subsection{Architectural Details}
In the proposed framework, we employ a transformer-based architecture (a pre-trained ViT-B/16 \cite{farl} consisting of 12 layers and a width of 768) as a backbone. To enhance its performance further, we incorporated three custom CNN-based hourglass modules after every four layers of the transformer network. The purpose of this module is to introduce desirable properties of CNNs, such as shift, scale, and distortion invariance, into the ViT architecture, while still retaining the characteristics of transformers, i.e., dynamic attention, global context, and better generalization. This results in a robust backbone network (Transformer + CNN) which learns facial structures effectively.

The utilization of pooling layers in CNNs often provides a certain degree of shift invariance in the model. However, in our task, it is imperative to avoid the loss of structural information caused by pooling layers. 
We therefore adopt the Anti-aliased CNN \cite{antialiasing} into our hourglass modules, hereafter known as Anti-aliased Hourglass. The combination of these components significantly enhances the caliber of our network towards high-quality heatmap generation. Nevertheless, the upsampling + concatenation (U+A) operation in the hourglass modules may introduce some high-frequency noise. To mitigate this negative impact and filter the features in the Fourier space, we integrate a FF-Parser layer \cite{ffparser} after each U+A operation in the hourglass modules. We provide ablation studies on these components in our results to demonstrate their usefulness.


\section{Experiments and Results}
\label{sec:exp}
This section discusses the implementation details, comparison with SOTA methods on benchmark datasets and ablation analysis of the introduced components of the proposed method. 

\vspace{4pt}
\noindent \textbf{Implementation Details:} The proposed method is trained/tested on the various benchmark datasets, i.e., WFLW \cite{WFLW}, 300W \cite{300W}, COFW \cite{COFW} and AFLW \cite{AFLW}. Details of these datasets are discussed in the Supplementary material. During the training phase, the input image is cropped and resized to $512 \times 512$. The output feature map size of every hourglass module is set to $128\times 128$, which is $4\times$ smaller than the input image size. The ground truth heatmaps are generated by a Gaussian with $\sigma = 1.5$ and radius $r = 5$. During training process, we used AdamW \cite{AdamW} to optimize our network with the initial learning rate of $1\times 10^{-4}$ and trained for 250 epochs. Apart from the proposed augmentation (\net), other standard data augmentations ($f_{A_1}$) are employed at training time, such as random masking, bilinear interpolation, random occlusion, random gray, random gamma, random blur, noise fusion. For effective learning, along with the DCCA loss (i.e., $\mathcal{L}_{DCCA}$), we also employ the standard BCE loss (i.e., $\mathcal{L}_{BCE}$) and mean absolute error loss (i.e., $\mathcal{L}_{L1}$) for heatmap and coordinate regression, respectively with equal weights (i.e., 1.0). For evaluation, we used the standard evaluation metrics i.e., Normalized Mean Error ($NME$) variants (i.e., $NME_{ic}$, $NME_{box}$, $NME_{diag}$), Failure Rate ($FR^{10}_{ic}$), Area Under the Curve ($AUC_{box}$). Detailed definitions of these metrics have been discussed in the Supplementary material. For comparison, we choose recent baselines such as FaRL \cite{farl}, ADNet \cite{adnet}, SH-FAN \cite{shfan}, PropNet \cite{propnet}, HIH \cite{hih}, SLPT \cite{slpt}, PicassoNet \cite{picassonet} and DTLD \cite{dtld}. All the experiments were implemented using PyTorch and the network was trained on 4 GPUs (40GB NVIDIA A100), with batch size 5 per GPU.

\vspace{-0.3em}
\subsection{Result Analysis}

\begin{table}[t!]
\centering
\caption{\small Comparison against the state-of-the-art on COFW, 300W and AFLW dataset. Best result is \textbf{bolded} and second best result is \underline{underlined}.}\label{table1}
\vspace{0.3em}
\begin{adjustbox}{width=.99\linewidth,center}
\begin{tabular}{|l|l|cc|ccc|cccc|}
\hline
\multicolumn{1}{|c|}{\multirow{3}{*}{Method}} & \multicolumn{1}{c|}{\multirow{3}{*}{Remarks}} & \multicolumn{2}{c|}{COFW}                        & \multicolumn{3}{c|}{300W}  & \multicolumn{4}{c|}{AFLW}                     \\ \cline{3-11} 
&   & \multirow{2}{*}{$NME_{ic}\downarrow$} & \multirow{2}{*}{$FR^{10}_{ic}\downarrow$} & \multicolumn{3}{c|}{$NME_{ic}\downarrow$} & \multicolumn{2}{c}{$NME_{diag}\downarrow$} & $NME_{box}\downarrow$ & $AUC_{box}\uparrow$ \\ \cline{5-11} 
&    &                        &                         & Full  & Common & Challenge & Full        & Frontal       & Full   & Full   \\ \hline
FaRL \cite{farl}      &        CVPR '22                            & 3.11                   & \underline{0.12}                    & \underline{2.93}  & 2.56   & 4.45      & \underline{0.94}        & \underline{0.82}          & \underline{1.33}   & \underline{81.3}   \\
ADNet \cite{adnet}     &      ICCV '21                              & 4.68                   & 0.59                    & \underline{2.93}  & \underline{2.53}   & 4.58      & ---         & ---           & ---    & ---    \\
SH-FAN \cite{shfan}     &       BMVC '21                           & \underline{3.02}       & \textbf{0.00}                    & 2.94  & 2.61   & \underline{4.13}      & 1.31        & 1.12          & 2.14   & 70.0   \\
PropNet \cite{propnet}   &        CVPR '20                           & 3.71                   & 0.20                    & \underline{2.93}  & 2.67   & \textbf{3.99}      & ---         & ---           & ---    & ---    \\
HIH \cite{hih}            &        ICCVW '21                      & 3.21                   & \textbf{0.00}                    & 3.09  & 2.65   & 4.89      & ---         & ---           & ---    & ---    \\
SLPT \cite{slpt}           &         CVPR '22                      & 3.32                   & 0.59                    & 3.17  & 2.75   & 4.90      & ---         & ---           & ---    & ---    \\
DTLD \cite{dtld}                 &         CVPR '22               & \underline{3.02}                   & ---                     & 2.96  & 2.60   & 4.48      & 1.37        & ---           & ---    & ---    \\
PicassoNet \cite{picassonet}    &         TNNLS '22                          & ---                    & ---                     & 3.58  & 3.03   & 5.81      & 1.59        & 1.30          & ---    & ---    \\
\rowcolor{LightCyan}
\emph{FiFA} (Ours)                    &       \multicolumn{1}{|c|}{---}         & \textbf{2.96}                   & \textbf{0.00}                    & \textbf{2.89}  & \textbf{2.51}   & 4.47      & \textbf{0.92}        & \textbf{0.80}          & \textbf{1.31}   & \textbf{81.8}   \\ \hline
\end{tabular}
\end{adjustbox}
\end{table}

\begin{table}
\centering
\caption{\small Comparison against the state-of-the-art on WFLW testset. Best result is \textbf{bolded} and second best result is \underline{underlined}.}\label{table:WFLW}
\vspace{0.3em}
\begin{adjustbox}{width=.99\linewidth,center}
\begin{tabular}{|l|l|l|c|cccccc|}
\hline
\multirow{2}{*}{Metric}  & \multirow{2}{*}{Models} & \multirow{2}{*}{Remarks}  & \multirow{2}{*}{Fullset} & \multicolumn{6}{c|}{Subset}   \\ \cline{5-10}  
&          &          &                         & Pose   & Expression & Illumination & Make Up & Occlusion & Blur\\ \hline
\multirow{9}{*}{$NME_{\text{ic}}$(\%)$\downarrow$} & FaRL \cite{farl}   &       CVPR'22                              & {3.99}                     & {\underline{6.61}}   & {4.18}       & {\underline{3.90}}         & {\underline{3.84}}    & {4.71}      & 4.53                \\                        & ADNet \cite{adnet}       &           ICCV'21                     & {4.14}                     & {6.96}   & {4.38}       & {4.09}         & {4.05}    & {5.06}      & 4.79   \\
                         & SH-FAN  \cite{shfan}  &           BMVC'21                        & {\textbf{3.72}}                     & {---}    & {---}        & {---}          & {---}     & {---}       & ---               \\
                         & PropNet \cite{propnet}  &          CVPR'20                          & {4.05}                     & {6.92}   & {3.87}       & {4.07}         & {\textbf{3.76}}    & {\textbf{4.58}}      & \textbf{4.36}       \\
                         & HIH \cite{hih}          &           ICCVW'21                    & {4.08}                     & {6.87}   & {\textbf{4.06}}       & {4.34}         & {3.85}    & {4.85}      & 4.66              \\
                         & SLPT \cite{slpt}         &            CVPR'22                   & {4.14}                     & {6.96}   & {4.45}       & {4.05}         & {4.00}    & {5.06}      & 4.79                \\
                         & DTLD \cite{dtld}            &       CVPR'22                     & {4.05}                     & {---}    & {---}        & {---}          & {---}     & {---}       & ---  \\
                         & PicassoNet \cite{picassonet}    &        TNNLS'22                      & {4.82}                     & {8.61}   & {5.14}       & {4.73}         & {4.68}    & {5.91}      & 5.56                \\
                         & \cellcolor{LightCyan} \emph{FiFA} (Ours)   &  \cellcolor{LightCyan}  ---                               & \cellcolor{LightCyan}\underline{3.89}                     & \cellcolor{LightCyan}\textbf{6.47}   & \cellcolor{LightCyan}\underline{4.09}       & \cellcolor{LightCyan}\textbf{3.80}         & \cellcolor{LightCyan}\textbf{3.76}    & \cellcolor{LightCyan}\underline{4.63}      & \cellcolor{LightCyan}\underline{4.43}  \\ \hline
\multirow{9}{*}{$FR_{\text{ic}}^{10}$(\%)$\downarrow$}  & FaRL \cite{farl}  &     CVPR'22                                 & {1.76}                     & {---}    & {---}        & {---}          & {---}     & {---}       & ---  \\
                         & ADNet \cite{adnet}         &          ICCV'21                    & {2.72}                     & {12.72}  & {\underline{2.15}}       & {2.44}         & {1.94}    & {5.79}      & 3.54  \\
                         & SH-FAN \cite{shfan}        &          BMVC'21                    & {\textbf{1.55}}                     & {---}    & {---}        & {---}          & {---}     & {---}       & ---   \\
                         & PropNet \cite{propnet}     &           CVPR'20                     & {2.96}                     & {\underline{12.58}}  & {2.55}       & {2.44}         & {\underline{1.46}}    & {\underline{5.16}}      & 3.75   \\
                         & HIH \cite{hih}           &           ICCVW'21                   & {2.60}                     & {12.88}  & {\textbf{1.27}}       & {2.43}         & {\textbf{1.45}}    & {\underline{5.16}}      & \underline{3.10}   \\
                         & SLPT \cite{slpt}        &           CVPR'22                     & {2.76}                     & {12.72}  & {2.23}       & {\underline{1.86} }        & {3.40}    & {5.98}      & 3.88  \\
                         & DTLD \cite{dtld}              &        CVPR'22                  & {2.68}                     & {---}    & {---}        & {---}          & {---}     & {---}       & ---  \\
                         & PicassoNet \cite{picassonet} &            TNNLS'22                      & {5.64}                     & {25.46}  & {5.10}       & {4.30}         & {5.34}    & {10.59}     & 7.12   \\
                         & \cellcolor{LightCyan} \emph{FiFA} (Ours)  &  \cellcolor{LightCyan} ---                                  & \cellcolor{LightCyan}\underline{1.60}                     & \cellcolor{LightCyan}\textbf{7.05}   & \cellcolor{LightCyan}\textbf{1.27}       & \cellcolor{LightCyan}\textbf{1.43}         & \cellcolor{LightCyan}\textbf{1.45}    & \cellcolor{LightCyan}\textbf{3.39}      & \cellcolor{LightCyan}\textbf{1.94} 
                         \\ \hline

\end{tabular}
\end{adjustbox}
\end{table}

\noindent \textbf{Comparison on COFW:} 
In Table \ref{table1}, we presents a comparison of the proposed \net\ approach with existing SOTA methods on the COFW testset, which is a well-known benchmark for heavy occlusion and a wide range of head pose variation. It is noteworthy that the proposed \net\ model outperforms the existing SOTA methods. The leading $NME_{ic}$ and $0\%$ $FR_{ic}^{10}$ demonstrate its robustness against extreme situations. 

\vspace{4pt}
\noindent \textbf{Comparison on 300W:} 
On the 300W dataset, our approach exhibits superior performance in comparison to SOTA methods in terms of $NME_{ic}$, and is given in Table \ref{table1}. In challenge-set, the proposed approach performs slightly lower than PropNet \cite{propnet} and SH-FAN \cite{shfan} methods. However, it has achieved SOTA results in other scenarios (i.e., full-set and common-set), which suggests that our method makes plausible predictions even in deplorable situations.

\vspace{4pt}
\noindent \textbf{Comparison on AFLW:} The results on AFLW testset are presented in Table \ref{table1}. Adhering to the evaluation protocol adopted in \cite{farl}, we report comparisons in terms of $NME_{diag}$, $NME_{box}$ and $AUC_{box}^7$. This table clearly indicates that our approach has outperformed the SOTA results, despite the fact that the dataset is almost saturated. 

\begin{figure}[t!]
    \centering
    \includegraphics[width = 0.99\linewidth]{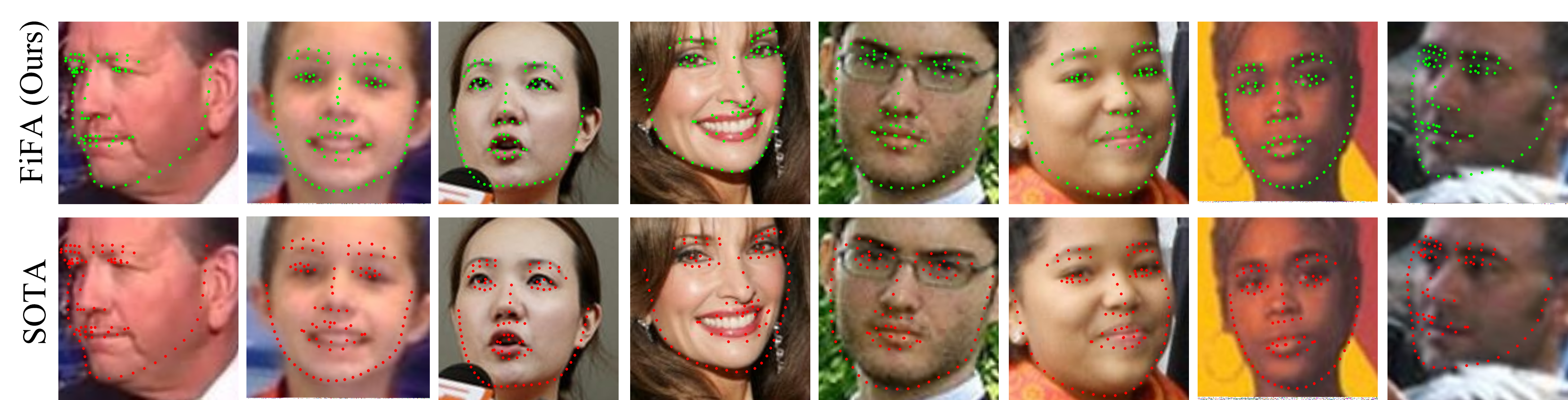}
    \caption{\small \textbf{Qualitative results on WFLW testset}. Landmarks shown in \textcolor{green}{green} are produced by our method, while the ones in \textcolor{red}{red} by the state-of-the-art approach of \cite{farl}. 
    }\label{fig:visual}
\end{figure}

\vspace{4pt}
\noindent \textbf{Comparison on WFLW:} 
In Table \ref{table:WFLW}, we compare results in terms of $NME_{ic}$, and $FR_{ic}^{10}$. Here, it is observed that the proposed \net\ approach obtains better $NME_{ic}$ for Pose, Illumination and Make Up subsets. Additionally, in comparison on $FR_{ic}^{10}$, the proposed approach achieves higher performance in all subsets i.e., Pose, Expression, Illumination, Make Up, Occlusion, Blur by 44\%, 41\%, 23\%, 1\%, 34\%, 37.4\%, respectively over the previous best performing SOTA methods. These results show that our method improves the accuracy in challenging scenarios while also reducing the overall failure ratio for difficult images. Moreover, Fig. \ref{fig:visual} visually conveys that the proposed approach delivers significantly more precise landmarks in challenging scenarios.

\subsection{Ablation Studies \& Analysis}
\label{sec:ablation}
This section presents the ablation analysis carried out to establish the efficacy of the proposed framework. To ensure fair comparison, all experiments were performed on COFW dataset.

\vspace{4pt}
\noindent \textbf{Effects of method's components:} 
Herein, we investigate the impact of each component of the proposed framework. The results, presented in Table \ref{table:components}, reveal that the baseline network, i.e., Vanilla backbone (ViT-B/16), attains an $NME_{ic}$ of 3.11 when trained solely with standard augmentations, i.e., $f_{A_1}$. When anti-aliased CNN-based hourglass modules are incorporated into baseline, an improvement in $NME_{ic}$ to 3.07 is observed. By employing the proposed augmentation, $f_{A_2}$, on the input images during training, a remarkable performance boost is achieved, with an $NME_{ic}$ of 3.00. The highest $NME_{ic}$ of 2.96 is attained when incorporating the Siamese training approach with DCCA loss on both $f_{A_1}$ and $f_{A_2}$ augmented images. This finding demonstrates that training the backbone with proposed components gives best performance in results.

\begin{table}[t]
\parbox{.49\linewidth}{
\caption{\small Effect of method's components on COFW.} \label{table:components}
\vspace{0.5em}
\begin{adjustbox}{width=.99\linewidth,center}
\begin{tabular}{|l|c|}
\hline
Method                        & $NME_{ic}$(\%)$\downarrow$ \\ \hline
Vanilla backbone (ViT-B/16)                     & 3.11    \\
+ anti-aliased CNN-based hourglass          & 3.07    \\
+ Fiducial Focus Augmentation & 3.00    \\
\rowcolor{LightCyan} + Siamese training (w DCCA)   & \textbf{2.96}    \\ \hline
\end{tabular}
\end{adjustbox}
\vspace{0.7cm}
\caption{\small Effect of patch sizes in \net\ on COFW.}\label{table:patch}
\vspace{0.5em}
\begin{adjustbox}{width=.95\linewidth,center}
\begin{tabular}{|c|c|}
\hline
\net\ patch progression & $NME_{ic}(\%)\downarrow$  \\ \hline
 3$\times$3 $\rightarrow \cdot \cdot \cdot \rightarrow 1\times1 \rightarrow$ no patch    & 3.05 \\
 4$\times$4 $\rightarrow \cdot \cdot \cdot \rightarrow 1\times1 \rightarrow$ no patch    & 3.00 \\
 \rowcolor{LightCyan} 5$\times$5 $\rightarrow \cdot \cdot \cdot \rightarrow 1\times1 \rightarrow$ no patch    & \textbf{2.96} \\
 6$\times$6 $\rightarrow \cdot \cdot \cdot \rightarrow 1\times1 \rightarrow$ no patch    & 2.99 \\
 7$\times$7 $\rightarrow \cdot \cdot \cdot \rightarrow 1\times1 \rightarrow$ no patch    & 3.02 \\ \hline
\end{tabular}
\end{adjustbox}
}
\hfill
\parbox{.49\linewidth}{
\caption{\small Effect of \net\ over standard augmentations on COFW. BI = Bilinear Interpolation; RM = Random Masking; RO = Random Occlusion; RGr = Random Gray; RGm = Random Gamma; RB = Random Blur; NF = noise fusion.}\label{table:fiFA effect}
\vspace{0.5em}
\begin{adjustbox}{width=.99\linewidth,center}
\begin{tabular}{|l|c|}
\hline 
Augmentations  & $NME_{ic}$(\%)$\downarrow$  \\ \hline
RM + RO  & 3.15 \\ 
\rowcolor{LightCyan} \quad + \net\  & 3.08 \\ \hline
RM + \{RO, RGr\}  & 3.12 \\
\rowcolor{LightCyan} \quad + \net\  & 3.07 \\ \hline
RM + \{RO, RGr, RGm\}  & 3.10 \\
\rowcolor{LightCyan} \quad + \net\ & 3.04 \\ \hline
RM + \{RO, RGr, RGm, RB\}   & 3.10 \\
\rowcolor{LightCyan} \quad + \net\ & 3.04 \\ \hline
RM + BI + \{RO, RGr, RGm, RB\} & 3.08 \\
\rowcolor{LightCyan} \quad + \net\  & 3.03 \\ \hline
RM + BI + NF + \{RO, RGr, RGm, RB\} & 3.07 \\
\rowcolor{LightCyan} \quad + \net\  & 3.00 \\ \hline
\end{tabular}
\end{adjustbox}
}
\vspace{-1em}
\end{table}

\vspace{4pt}
\noindent \textbf{Effects of fiducial mask sizes:} 
We have conducted a series of experiments to determine the optimal initial patch size for the proposed \net. As shown in Table \ref{table:patch}, a patch size of $5\times 5$ yields the best $NME_{ic}$ of 2.96, while deviating from this size leads to a deterioration in performance. This can be attributed to the fact that during the initial stages of training, when the network weights are not yet sufficiently tuned, a patch size that is either too large or too small will result in a confidence region that is either too broad or too narrow for the network to focus on the landmarks. This, in turn, has an adverse effect on the learning process and ultimately on the performance of the network.

\vspace{4pt}
\noindent \textbf{Effect of \net\ over standard augmentations:} 
Several experiments were conducted to prove the effectiveness of our proposed \net\ over other standard augmentations. Due to the availability of only one view of augmented images, all these experiments were performed without a Siamese-based training mechanism. Table \ref{table:fiFA effect} displays the results obtained in terms of $NME_{ic}$ on the COFW testset. One can notice that the inclusion of our proposed \net\ in standard augmentation techniques leads to a notable improvement in the $NME_{ic}$ value.

\vspace{4pt}
\noindent \textbf{Comparison with other losses in Siamese training:} 
We employ DCCA loss \cite{dcca} in Siamese training to maximize the correlation between different views. To demonstrate the efficacy of DCCA loss, we conducted several experiments with different losses (i.e., L2, L1, Smooth L1, and Wing loss \cite{wing}), and the corresponding results are presented in Table \ref{table:loss}. One can observe that the DCCA loss helps to obtain better $NME_{ic}$, exhibiting a 3\% increase as compared to previous best-performing Wing loss.
\begin{table}[h!]
\caption{\small Effect of different losses in Siamese training on COFW.} \label{table:loss}
\vspace{0.5em}
\begin{adjustbox}{width=.69\linewidth,center}
\begin{tabular}{|l|ccccc|}
\hline
Loss    & L2   & L1   & Smooth L1 & Wing \cite{wing} & \cellcolor{LightCyan} DCCA \cite{dcca} \\ \hline
$NME_{ic}$(\%)$\downarrow$ & 3.14 & 3.09 & 3.11 & 3.05 & \cellcolor{LightCyan} \textbf{2.96} \\ \hline
\end{tabular}
\end{adjustbox}
\end{table}

\noindent\textbf{Effectiveness of the proposed components to other SOTA methods:} To validate the effectiveness of the proposed components, we conducted a series of experiments wherein the proposed \net\ augmentation and Siamese network based DCCA loss were implemented on other baseline methods such as HRNet \cite{wang2020deep}, ADNet \cite{adnet}, SH-FAN backbone \cite{shfan}, FaRL \cite{farl}, SLPT \cite{slpt} and the corresponding results are summarized in Table \ref{table:baseline}. The proposed \net\ augmentation technique improved the performance of baseline methods. Additionally, the Siamese network based DCCA loss contributed to improve the NME score further. This clearly indicates the generalization capability of our method.
\begin{table}[h!]
\caption{\small Effect of proposed \net\ augmentation technique and Siamese-based DCCA loss on baseline methods on COFW testset.} \label{table:baseline}
\vspace{0.5em}
\begin{adjustbox}{width=0.85\linewidth,center}
\begin{tabular}{|l|l|c|c|c|}
\hline
\begin{tabular}[c]{@{}c@{}}Methods\end{tabular} & \multicolumn{1}{c|}{Remarks} & Baseline & + \net\ & \cellcolor{LightCyan} \begin{tabular}[c]{@{}c@{}}+ \net\ \\ + Siamese training (w DCCA)\end{tabular} \\ \hline
HRNet \cite{wang2020deep}                              & $ICCV_{21}$                      & 3.45            &   3.32    & \cellcolor{LightCyan} \textbf{3.28}                      \\
ADNet \cite{adnet}                              & $ICCV_{21}$                      & 4.68            & 4.51      & \cellcolor{LightCyan} \textbf{4.45}                    \\
SH-FAN Backbone \cite{shfan}     & $BMVC_{21}$   & 3.25            & 3.12      & \cellcolor{LightCyan} \textbf{3.07}                    \\ 
FaRL \cite{farl}                               & $CVPR_{22}$                      & 3.11            & 3.04      & \cellcolor{LightCyan} \textbf{3.01}                    \\
SLPT \cite{slpt}                               & $CVPR_{22}$                      & 3.32            & 3.15      & \cellcolor{LightCyan} \textbf{3.10}                    \\ 
\hline
\end{tabular}
\end{adjustbox}
\end{table}

\section{Conclusion \& Future Work}\label{sec:conclusion}
In this paper, we successfully proposed a simple yet effective image augmentation technique called Fiducial Focus Augmentation (\net) for facial landmark detection task. 
The integration of \net\ during training significantly enhanced the accuracy of proposed approach on testing benchmarks without extreme modifications to its backbone network and the loss function. Our findings suggest that the employment of \net\ as an image augmentation technique, when used in conjunction with a Siamese-based training with DCCA loss results in state-of-the-art performance. 
Additionally, we employed an anti-aliased CNN-based hourglass network with ViT as our backbone network to address shift invariance and noise. We performed extensive experimentation and ablation studies to validate the effectiveness of the proposed approach. In future work, \net\ can be studied further to extend it for other face-related tasks.


\bibliography{egbib.bib}
\end{document}